\documentclass[titlepage,11pt]{article}

\usepackage[pdftex]{graphicx}

\usepackage[myheadings]{fullpage}
\usepackage{pmetrika}
\usepackage{pmbib}
\usepackage{apacite}

\usepackage{submit}

\usepackage{amsmath}

\newcommand{\rv}[1]{\underline{#1}}

\newcommand{\ve}[1]{{\bf {#1}}}
\newcommand{\term}[1]{\textit{#1}}
\newcommand{\ex}{{\rm E}}
\newcommand{\andOp}{{\rm and}}
\newcommand{\orOp}{{\rm or}}
\newcommand{\pickOp}{{\rm pick}}
\newcommand{\partOp}{{\rm part}}

\newcommand{\paperTitle}{Tracking student skills real-time through a continuous-variable dynamic Bayesian network}

\begin{document}

\begin{titlepage}

\title{\paperTitle}


\author{Hildo Bijl}

\affil{}


\vspace{\fill}\centerline{August 19, 2024}\vspace{\fill}

\linespacing{1}



\end{titlepage}

\setcounter{page}{2}
\vspace*{2\baselineskip}

\RepeatTitle{\paperTitle}\vskip3pt

\linespacing{1.5}
\abstracthead
\begin{abstract}
The field of Knowledge Tracing is focused on predicting the success rate of a student for a given skill. Modern methods like Deep Knowledge Tracing provide accurate estimates given enough data, but being based on neural networks they struggle to explain how these estimates are formed. More classical methods like Dynamic Bayesian Networks can do this, but they cannot give data on the accuracy of their estimates and often struggle to incorporate new observations in real-time due to their high computational load. 

This paper presents a novel method, Performance Distribution Tracing (PDT), in which the distribution of the success rate is traced live. It uses a Dynamic Bayesian Network with continuous random variables as nodes. By tracing the success rate distribution, there is always data available on the accuracy of any success rate estimation. In addition, it makes it possible to combine data from similar/related skills to come up with a more informed estimate of success rates. This makes it possible to predict exercise success rates, providing both explainability and an accuracy indication, even when an exercise requires a combination of different skills to solve. And through the use of the beta distribution functions as conjugate priors, all distributions are available in analytical form, allowing efficient online updates upon new observations. Experiments have shown that the resulting estimates generally feel sufficiently accurate to end-users such that they accept recommendations based on them.

\end{abstract}\vspace{\fill}\pagebreak

\section{Introduction}

When a teacher/tutor interacts with a student, the teacher tracks (often subconsciously) the student's progress. There is always an idea in the back of the teacher's mind, `The student is definitely good with these skills and probably less good with these other skills.' This tracking is subject to four important properties.
\vspace{-6pt}
\begin{enumerate}
    \itemsep -2pt
    \item \label{req:realTime} Success rate estimates are done real-time, taking into account a student's learning progress.
    \item \label{req:originAccuracy} The teacher can clarify the origin as well as the certainty/accuracy of said estimates.
    \item \label{req:domainKnowledge} The estimates take into account links between a large variety of skills.
\end{enumerate}
\vspace{-6pt}
The goal of this paper is to set up an algorithm that satisfies the above four requirements. The resulting success rate estimates can then be used to (among others) provide recommendations to students on what skills to practice and which exercises are on the right level to do so. This effectively automates a large part of what an individual tutor does, but then in a way that is accessible to a large amount of (possibly disadvantaged) students, and guaranteed without any potential subconscious bias. The final goal of this paper is that the resulting skill level estimates feel sufficiently realistic to the end-users, so that they will accept recommendations based on them just as they would for teacher recommendations.

There is a long history behind skill tracking and digital tutoring systems. Most work relates to \term{Knowledge Components} (KCs), as described by~\cite{KnowledgeComponents}. A distinction can be made between \term{explicit KCs} (facts/principles) and \term{implicit KCs} (skills). Skills are generally practiced and tested through exercises, and the fundamental question then is, `What is the chance that the student will do the next exercise correctly?' This probability is known as the \term{success rate} and the study of tracking/estimating it is known as \term{Knowledge Tracing} (KT). Overviews are given by~\cite{KTModelsComparison} and~\cite{KTOverview} though a brief summary is provided below.

Classical work on KT revolves around \term{Bayesian Knowledge Tracing} (BKT)~\cite{BayesianKnowledgeTracing}. This method directly estimates the success rate, constantly updating this estimate on each subsequent exercise execution using a hidden Markov model. Probabilities of the student slipping (failing despite having mastery) or guessing (succeeding while not having mastery) are taken into account~\cite{BKTSlipAndGuess} as well as potentially the exercise difficulty~\cite{BKTExerciseDifficulty}. Extensions exist for further individualization towards students~\cite{BKTIndividualization1,BKTIndividualization2}.

Instead of using models that directly have probabilities as parameters, an alternative set of methods uses logistic models to estimate success rates. The fundamentals have been presented by~\cite{LFABasis} in the \term{Learning Factors Analysis} (LFA) method. By using weight factors to account for the difficulty of a KC, the ease of learning of a KC and various other parameters, and by inserting the result into a logistic function, the success rate for a KC is estimated. Extensions/variants include \term{Performance Factor Analysis} (PFA)~\cite{PFABasis} and \term{Knowledge Tracing Machines} (KTM)~\cite{KTMBasis}.

The methods above mainly work for individual KCs, struggling to take into account links between KCs. This so-called \term{domain knowledge} can be taken into account in two ways: prespecified or learned by the algorithm. With the introduction and growing popularity of \term{Massive Open Online Courses} (MOOCs)~\cite{MOOCs} the amount of data that can be used for KT has increased significantly. As a result, more advanced methods like \term{Deep Knowledge Tracing} (DKT) have appeared, with fundamental work done by~\cite{DKTBasis}. In DKT recurrent neural networks are used to model students' performance as they work through a course. Though this method requires a large amount of data, and is prone to overfitting~\cite{ATKT}, it does not require domain knowledge: the algorithm determines for itself which links between KCs exist. Possible extensions include taking into account the learning curve~\cite{DKTLearningCurves}. A downside is that neural networks struggle to explain their estimates and their accuracy, while such accountability is becoming increasingly more important in machine learning applications~\cite{MLAccountability}. Hence, to satisfy requirement~\ref{req:originAccuracy}, it is generally better to directly use existing domain knowledge, especially since such domain knowledge in education is readily available.

One field of study that takes into account domain knowledge is that of Cognitive Diagnostic Models (CDMs). This is a rather broad field, of which an overview is given by~\cite{CDMOverview}. The main idea is once more to model student performance by estimating the chance a student will do a certain exercise correctly, very similarly to KT. The strength of CDMs is that it tracks mastery for a variety of skills, and can then predict performance for exercises requiring multiple separate skills together. There are many interesting applications, for instance by~\cite{CMDwithHMM,CDMwithBN}. The downside of CDMs is that mastery is generally considered a binary variable -- it has been obtained or not -- so no data is available on the accuracy of the given estimates, violating requirement~\ref{req:originAccuracy}. It is also limited in how it can take into account links between skills, generally only considering exercises that require multiple skills all to be completed successfully (an and-operator), thus only partly satisfying requirement~\ref{req:domainKnowledge}.

To more specifically take into account domain knowledge, a \term{Bayesian Network} could be applied. When a learning effect should be considered, the network is then extended to a \term{Dynamic Bayesian Network} (DBN) as outlined by~\cite{BayesianPsychometrics}. There are various example applications of DBNs to education~\cite{DBKTBasis,DBNsInEducation,DBNInGameBasedAssessments1,DBNInGameBasedAssessments2,DBNSurvivalAnalysis}. Though giving very promising results, many of these implementations require numerically complex evaluations, often through methods like Gibbs sampling. This prevents the real-time updating of success rate estimates (requirement~\ref{req:realTime}). On top of this, all considered DBNs model skill acquisition as binary variables: there either is or isn't mastery. A claim `There is 60\% chance of mastery' hence does not have any accuracy data attached to it. Do we simply not know much about the student and hence loosely guess `60\%' as a prior guess? Or has an extensive set of data shown that the student successfully demonstrates a skill exactly 60\% of the time? (See Figure~\ref{fig:ExampleDistributions} later on for a more visual example of this.) This is left unclear, violating requirement~\ref{req:originAccuracy}.

In this paper a new method is proposed called \term{Performance Distribution Tracing} (PDT). In PDT the nodes of the DBN are not binary variables (mastery or not) whose chance is estimated, but continuous random variables (the success rates) whose distribution is constantly updated. By using beta distribution functions as conjugate priors, it is possible to update the success rate distributions analytically in real-time, satisfying requirement~\ref{req:realTime}. Since the \textit{distributions} of the success rates are known, it gives information on the certainty of the resulting success rate estimates (requirement~\ref{req:originAccuracy}). And knowing the success rate distributions also allows combining data from multiple sources -- for instance multiple KCs/skills -- improving explainability (again requirement~\ref{req:originAccuracy}) and facilitating the use of domain knowledge (requirement~\ref{req:domainKnowledge}).

In this paper we walk through the presented algorithm as follows. In the first section we study tracing the success rate distribution of a student for only a single skill. The next section extends these methods to take into account exercises involving multiple skills. We go a step further in the third section, studying links between various skills and how these links can be taken into account to get more informed success rate estimates. Finally, the paper closes off with conclusions and recommendations.

\section{Modeling the proficiency of a single skill}\label{s:ModelingSingleSkill}

Consider any single skill $A$ (an implicit knowledge component) that a student might do. This skill $A$ could be as basic as `multiply two small numbers' or as advanced as `perform a complex engineering mechanics calculation'. For this skill, we want to track the student's performance, incorporating new observations and making predictions on future executions.

\subsection{Mathematically describing the probability of success}\label{ss:BasisFunctions}

Key to modeling the student's proficiency at skill $A$ is the \term{success rate}: the chance $a$ that a student performs it successfully. Existing KT methods see $a$ as a deterministic variable that must be found. However, this provides no data on the confidence with which $a$ is known: is it a rough guess or a near-certain estimate?

\begin{figure}[hbt]
\centering
\includegraphics[width=0.75\textwidth]{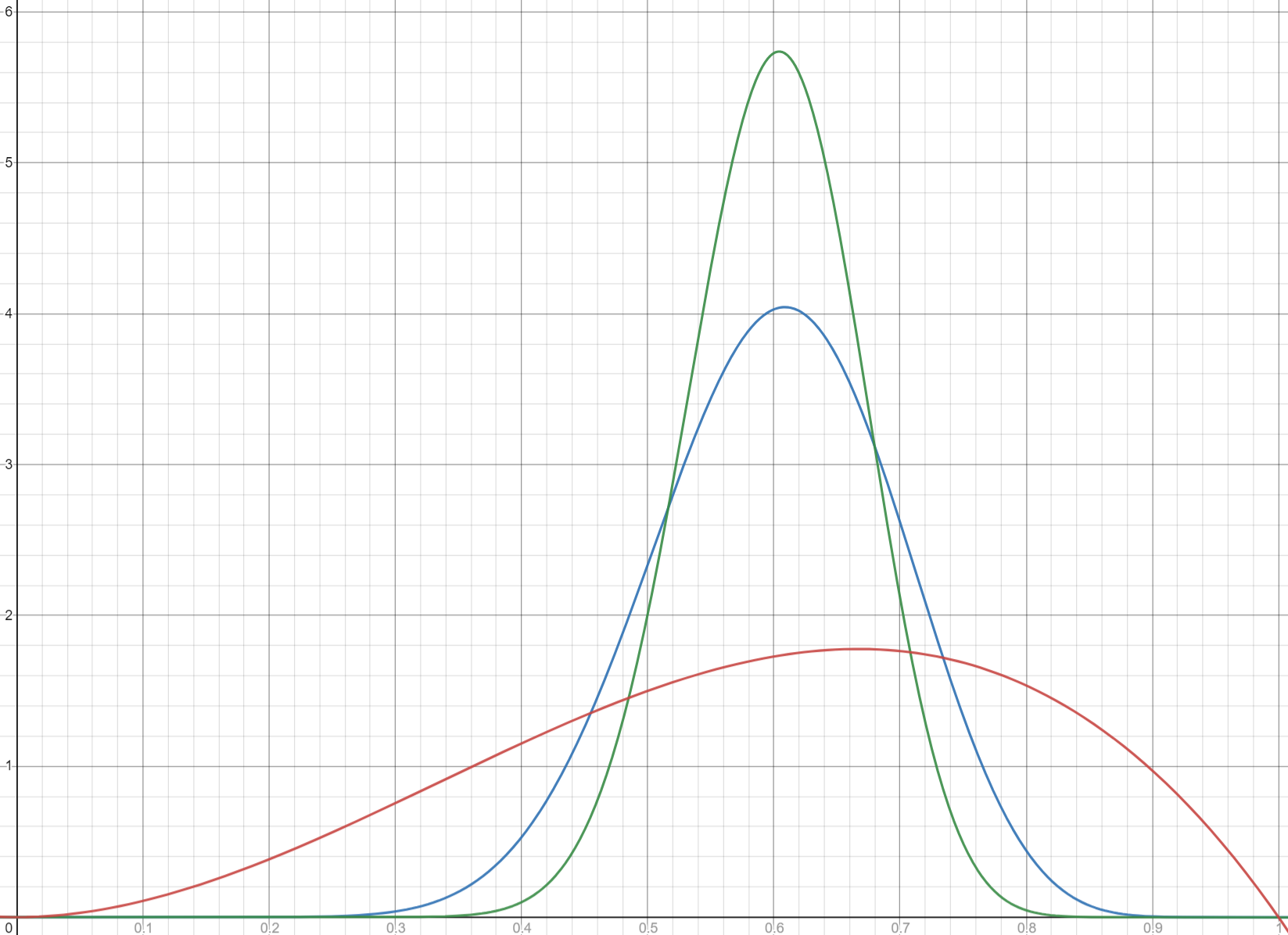}
\caption{Three example distributions $f_{\rv{a}}(a)$ plotted from $0$ to $1$. They all have an expected value of $0.6$, denoting a $60\%$ estimated success rate, but their peakedness and hence their degree of certainty varies. (The functions correspond to the situations [red: 2 correct, 1 incorrect; $n = 3, c_2 = 1$], [blue: 14 correct, 9 incorrect; $n = 23, c_{14} = 1$] and [green: 29 correct, 19 incorrect; $n = 48, c_{29} = 1$] with no learning effect or links taken into account.)}
\label{fig:ExampleDistributions}
\end{figure}

This problem is solved if we describe the success rate by a random variable $\rv{a}$. (The underline notation denotes random variables.) In this case $\rv{a}$ is not described by a single value, but by its distribution, expressed through its Probability Density Function (PDF) $f_{\rv{a}}(a)$. Examples of success rate PDFs are shown in Figure~\ref{fig:ExampleDistributions}. Note that $\rv{a}$ only takes values between $0$ and $1$ because it is a probability. We hence always have $f_{\rv{a}}(a) = 0$ for $a < 0$ or $a > 1$.

Describing the PDF of a success rate $\rv{a}$ is done through basis functions. Specifically, we use the probability density functions of the beta distribution~\cite[chapter~25]{BetaDistribution} as basis functions. We define the \term{basis functions} $g_{i,n}(a)$, for some \term{order} $n$, as
\begin{equation}\label{eq:BasisFunctionDefinition}
    g_{i,n}(a) = (n+1) \binom{n}{i} a^i \left(1-a\right)^{n-i}.
\end{equation}
Note that the above function is the PDF of a beta distribution with parameters $\alpha$ and $\beta$ satisfying $\alpha = i + 1$ and $\beta = n - i + 1$. Using the above function as basis function, we write the PDF of $\rv{a}$ as
\begin{equation}\label{eq:PDFWithCoefficientsAndBasisFunctions}
    f_{\rv{a}}(a) = \sum_{i=0}^n c_i g_{i,n}(a) = \ve{c}^T \ve{g_n}(a).
\end{equation}
Note that the \term{coefficient vector} $\ve{c}$ is the vector with all \term{coefficients} $c_i$ and identically the \term{basis function vector} $\ve{g_n}(a)$ is the vector function containing all basis functions $g_{i,n}(a)$. Obviously, not all functions can be described through this basis form, but by choosing our priors in the right way (as conjugate priors) the PDFs we encounter do fit this form. It effectively allows us to reduce a complete function to a set of coefficients.

There are some interesting consequences to describing the distribution of $\rv{a}$ in this way. Because $f_{\rv{a}}(a)$ is a PDF, its integral must equal one. Hence,
\begin{equation}
    \int_0^1 f_{\rv{a}}(a) \, da = \int_0^1 \left( \sum_{i=0}^n c_i g_{i,n}(a) \right) \, da = \sum_{i=0}^n c_i \left(\int_0^1 g_{i,n}(a) \, da\right) = \sum_{i=0}^n c_i = 1.
\end{equation}
Note that we have used the fact that the integral over the beta distribution PDF $g_{i,n}(a)$ equals 1. The above shows that the sum of all the coefficients must always equal one, and that there are hence only $n$ degrees of freedom in the $n+1$ coefficients. If the coefficients add up to one, the coefficients are said to be \term{normalized}. If the coefficients are not normalized, then it is always possible to normalize them by dividing each coefficient by the sum of all coefficients. In practice this is often done, as well as ensure that all coefficients remain non-negative, to compensate for potential numerical inaccuracies during computations.

Given $f_{\rv{a}}(a)$ we can also calculate the expected probability of success. This is given by
\begin{equation}\label{eq:ExpectedValue}
    \ex[\rv{a}] = \sum_{i=0}^n c_i \left(\int_0^1 a g_{i,n}(a) \, da\right) = \sum_{i=0}^n c_i \frac{i+1}{n+2}.
\end{equation}
More generally, any moment $\ex[\rv{a}^m]$ can be calculated through
\begin{equation}\label{eq:Moments}
    \ex[\rv{a}^m] = \sum_{i=0}^n c_i \frac{\left(n+1\right)!}{\left(n+m+1\right)!} \frac{\left(i+m\right)!}{i!}.
\end{equation}
Instead of looking at the success rate $\rv{a}$, we can also study the \term{failure rate} $1-\rv{a}$. The distribution of the failure rate, described by $f_{1-\rv{a}}(a)$, can be found by reversing the order of the coefficients $c_i$. That is,
\begin{equation}
    f_{1-\rv{a}}(a) = \sum_{i=0}^n c_{n-i} g_{i,n}(a).
\end{equation}
Or instead of looking at the PDF, we can also determine the CDF. Just like the PDF, the CDF is a linear combination of basis functions $g_{i,n+1}(a)$. Given a random variable $\rv{a}$, the CDF $F_{\rv{a}}(a)$ on the interval $0 \leq a \leq 1$ equals
\begin{equation}
    F_{\rv{a}}(a) = \int_0^a f_{\rv{a}}(x) \, dx = \sum_{i=0}^{n+1} C_i g_{i,n+1}(a) = \ve{C}^T \ve{g_{n+1}}(a), \hspace{10pt} \mbox{where} \hspace{10pt} C_i = \sum_{j=0}^{i-1} c_j.
\end{equation}
Note that the $(n+2)$ coefficients $C_i$ of the CDF are the cumulative coefficients of the $(n+1)$ coefficients $c_i$ of the PDF, starting with an extra zero coefficient. Also note that, outside of the interval $[0,1]$, we of course have $F_{\rv{a}}(a) = 0$ for $a < 0$ and $F_{\rv{a}}(a) = 1$ for $a > 1$.

This concludes the relevant properties of the basis function description. Let's apply them.

\subsection{Updating the skill success rate distribution}\label{ss:UpdatingSkillsBasic}

Initially, when no data is present on a skill $A$, we generally start with the flat prior $f_{\rv{a}}(a) = 1$. This corresponds to a coefficient vector of $\ve{c} = [1]$. The order of this coefficient vector is hence $0$: there is no information yet.

Next suppose that, after a student has done $k-1$ exercises for some skill $A$, the performance distribution $f_{\rv{a}}(a|D_{k-1})$ is known. Here, $D_{k-1}$ denotes all data related to the first $k-1$ exercise executions. In other words, the coefficient vector $\ve{c}$ is known up to this point. If the student then performs an exercise again (execution $k$), which is either a success or a failure, how is this data incorporated?

The posterior distribution is given through Bayes' law as
\begin{equation}\label{eq:UpdateLawBayesForm}
    f_{\rv{a}}(a|D_k) = \frac{p(D_k|a,D_{k-1}) f_{\rv{a}}(a|D_{k-1})}{p(D_k|D_{k-1})}.
\end{equation}
Note that, if the last exercise execution was successful, then $p(D_k|a,D_{k-1}) = a$, while on a failure it equals $p(D_k|a,D_{k-1}) = 1-a$. Also note that $p(D_k|D_{k-1})$ is a constant, not depending on $a$. Assuming that afterwards we normalize our coefficient vector $\ve{c}$, which we always do, we can safely ignore it.

The posterior distribution $f_{\rv{a}}(a|D_k)$ described above is once more a distribution in basis form. That means we can describe it through a new set of coefficients. If the old distribution has order $n$, the new distribution can be shown to have order $n_* = n+1$ with coefficients (for $0 \leq i \leq n_*$)
\begin{equation}\label{eq:BasicUpdateLaw}
    c_i^* = \begin{cases}
    i c_{i-1} & {\rm on\ success,} \\
    (n+1-i) c_i & {\rm on\ failing,}
    \end{cases}
\end{equation}
where afterwards we apply normalization to the coefficients to get their sum to equal one once more. (The full derivation of the above is omitted for reasons of brevity, but follows directly from~\eqref{eq:UpdateLawBayesForm}.) Note that $\ve{c^*}$ (the star superscript) denotes the coefficients of the \textit{updated} distribution, incorporating the latest observation. This star-notation will be used more often in this paper for the updated distribution. Through this update law, we can continuously incorporate more data, allowing the performance distribution to become more peaked over time. See for instance once more Figure~\ref{fig:ExampleDistributions}.

\subsection{Inferring the success rate for the next execution}\label{ss:Smoothing}

The above has assumed that the actual value of the success rate $\rv{a}$ is a constant: it may be unknown, but it does not vary over time. In reality this is not the case. After every exercise the student may have learned something new, or possibly misinterpreted something. The value of $\rv{a}$ hence slowly shifts, upwards or downwards. This is known as the \term{learning effect}~\cite{LearningEffect}.

To account for the learning effect, we use (as is customary in KT) multiple sequential random variables: we define $\rv{a}_k$ as the probability of success of execution $k$ of skill $A$. Initially we know (as our prior) the distribution $f_{\rv{a}_1}(a)$. After the first exercise, this will be updated using~\eqref{eq:BasicUpdateLaw} as $f_{\rv{a}_1}(a|D_1)$. This shows us the posterior probability that the student will execute the given skill successfully on the first try. But what does this tell us about the probability of success $\rv{a}_2$ of the second execution?

\begin{figure}[!ht]
\centering
\includegraphics[width=0.5\textwidth]{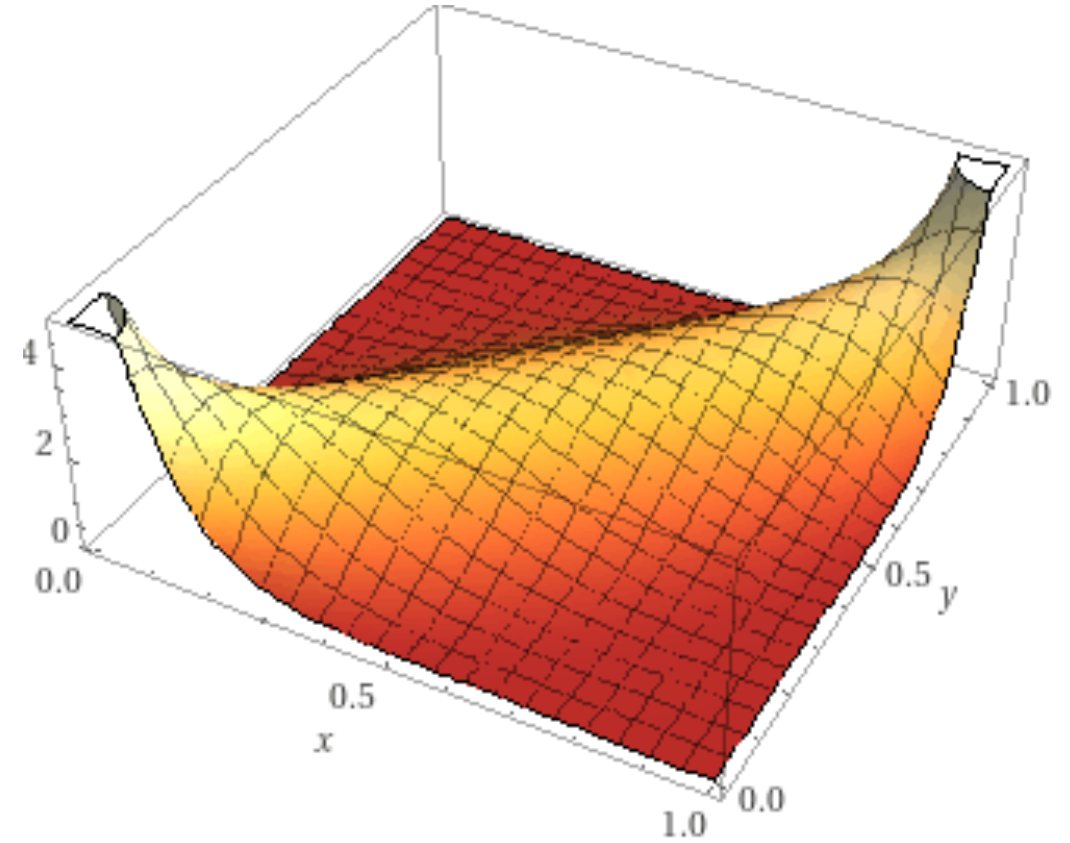}
\caption{The joint prior for two subsequent skill success rates $\rv{a}_k$ and $\rv{a}_{k+1}$, for smoothing order $n_s=10$. Note that, given a value of $\rv{a}_k$, the value of $\rv{a}_{k+1}$ will very likely be similar. Higher smoothing orders $n_s$ give a more peaked ridge at the diagonal line, and hence assume a stronger correlation between $\rv{a}_k$ and $\rv{a}_{k+1}$.}
\label{fig:JointPrior}
\end{figure}

To be able to say anything about $\rv{a}_2$, we must first assume a joint prior between subsequent skill executions $\rv{a}_k$ and $\rv{a}_{k+1}$: how similar are their success rates? Normally in KT, when the success rates $\rv{a}$ are modeled as binary random variables, this is done through a simple probability table. Instead, because we have continuous random variables, we must set up a joint prior probability density function. A convenient conjugate joint prior is
\begin{equation}\label{eq:JointPrior}
    f_{\rv{a}_k,\rv{a}_{k+1}}(a_k,a_{k+1}) = \frac{\ve{g_{n_s}}\hspace{-6pt}^T(a_k) \ve{g_{n_s}}(a_{k+1})}{n_s+1},
\end{equation}
where $n_s$ is known as the \term{smoothing order}. (This name will be explained later.) The shape of this prior is shown in Figure~\ref{fig:JointPrior} and the corresponding correlation between $\rv{a}_k$ and $\rv{a}_{k+1}$ is $\rho = \frac{n_s}{n_s+2}$. Note that the shape is symmetric with respect to $\rv{a}_k$ and $\rv{a}_{k+1}$. It may be sensible to skew this prior to encourage $\rv{a}_{k+1}$ to be (on average) larger than $\rv{a}_k$, since students generally get better at a skill with more practice instead of worse. However, for simplicity this has not been done. It could be a subject of future research.

The above prior can be used to infer the distribution of $\rv{a}_{k+1}$ from the distribution of $\rv{a}_k$. For this, we must first use the definition of the conditional distribution to write
\begin{equation}
    f_{\rv{a}_k,\rv{a}_{k+1}}(a_k,a_{k+1}|D_k) = f_{\rv{a}_{k+1}}(a_{k+1}|a_k,D_k) f_{\rv{a}_k}(a_k|D_k).
\end{equation}
Note that, given $a_k$, the data $D_k$ does not add any information to $\rv{a}_{k+1}$, so it can be omitted. This turns $f_{\rv{a}_{k+1}}(a_{k+1}|a_k,D_k)$ into $f_{\rv{a}_{k+1}}(a_{k+1}|a_k)$. Through the definition of the conditional probability, and by subsequently marginalizing over $\rv{a}_k$, we find
\begin{equation}\label{eq:SmoothingPDFs}
    f_{\rv{a}_{k+1}}(a_{k+1}|D_k) = \int_0^1 f_{\rv{a}_k,\rv{a}_{k+1}}(a_k,a_{k+1}|D_k) \, da_k = \int_0^1 \frac{f_{\rv{a}_k,\rv{a}_{k+1}}(a_k,a_{k+1}) f_{\rv{a}_k}(a_k|D_k)}{f_{\rv{a}_k}(a_k)} \, da_k.
\end{equation}
The resulting distribution is once more one in basis form. We can hence solve the integral and derive the resulting coefficients $c_{k+1,i}$. For the mathematically curious, the full derivation is given in Appendix~A. We only present the final outcome here as
\begin{equation}\label{eq:SmoothingCoefficients}
    c_{k+1,i} = \sum_{j=0}^{n_*} \binom{i+j}{i} \binom{n_* + n_s - i - j}{n_* - j} c_{k,j}^*,
\end{equation}
where afterwards the coefficients must be normalized as usual. Note that the coefficients $c_{k,j}^*$ describe $f_{\rv{a}_k}(a_k|D_k)$ (the success rate of execution $k$ \textit{given} the data on the result of execution $k$) while $c_{k+1,i}$ describe $f_{\rv{a}_{k+1}}(a_{k+1}|D_k)$ (the success rate of execution $k+1$ \textit{without} knowing the result of execution $k+1$). Also note that the order of this new set of coefficients $c_{k+1,i}$ is always the smoothing order $n_s$.

\begin{figure}[hbt]
\centering
\includegraphics[width=0.75\textwidth]{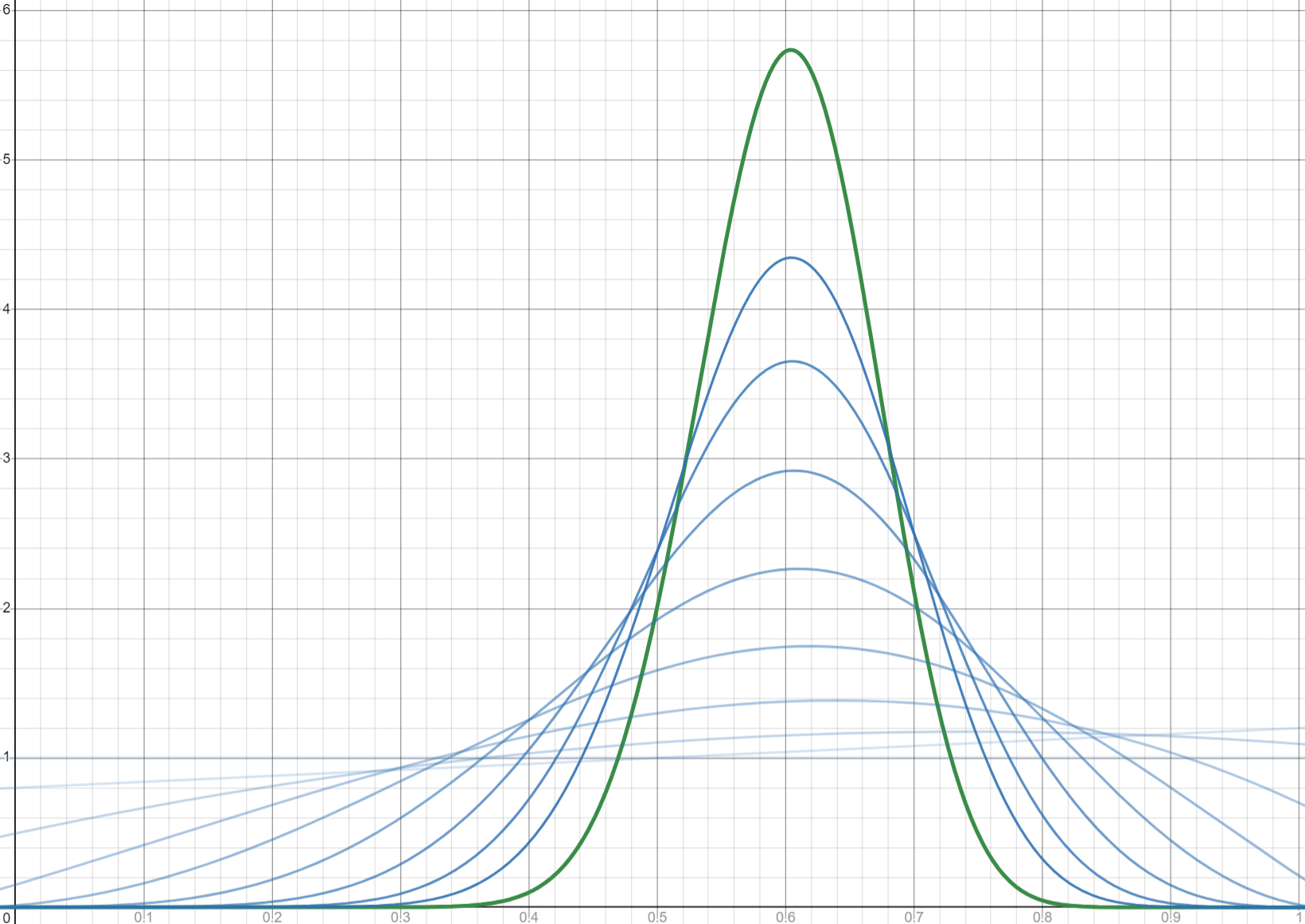}
\caption{An example distribution (green, $n = 48, c_{29} = 1$) that is smoothed through orders $128$, $64$, $32$, $16$, $8$, $4$, $2$ and $1$ (top to bottom) respectively. Higher smoothing orders leave the function mostly unchanged, while lower smoothing orders `flatten' it more. In fact, a smoothing order of $1$ guarantees a straight line, and a smoothing order of $0$ results in the flat prior $f_{\rv{a}}(a) = 1$. This smoothing is used to introduce extra uncertainty into the distribution, for instance due to the learning effect.}
\label{fig:SmoothingExample}
\end{figure}

When applying the joint prior in this way, the distribution of $\rv{a}_{k+1}$ is always slightly flatter (less peaked) than the distribution of $\rv{a}_k$. As a result, this step is known as the \term{smoothing step}. By applying this smoothing, we effectively incorporate the uncertainty of the learning effect. The smaller $n_s$ is, the more smoothing is applied, while a large value of $n_s$ leaves the distribution nearly unaffected. See for instance Figure~\ref{fig:SmoothingExample} for an example of smoothing a distribution.

The only unanswered question is how to choose $n_s$. In practice this is done based on various heuristic settings. The full method is discussed in Appendix~B but it can be summarized in a few simple rules. If a student has practiced the skill a lot, then the learning effect is smaller so $n_s$ is chosen to be larger (in the range $60-120$) to have less smoothing. Similarly, if the time since the last skill execution is large, the student may have forgotten a lot, so $n_s$ is chosen to be smaller (in the range $20-40$) to have more smoothing. In practice, because the smoothing step is time-dependent, the coefficients that are stored in the database for each student are always the coefficients $\ve{c_k^*}$ \textit{after} incorporating new observations yet \textit{prior} to smoothing.

\section{Handling exercises combining multiple skills}\label{s:Exercises}

So far we have considered exercises in which only a single skill needed to be applied. In practice, many exercises require a variety of skills. How can we incorporate results of these exercises into the corresponding skill success rate distributions? And how do we use this to predict exercise success rates?

\subsection{Mathematically defining the exercise set-up}\label{ss:ExerciseSetup}

The first step is to describe the exercise. Consider an exercise $X$ requiring skills $A$ and $B$. The exact way this exercise is set up can be described through various operators.
\begin{itemize}
    \item The $\andOp$-operator: if a successful execution of the exercise requires a successful execution of both $A$ and $B$, then we write $X = \andOp(A,B)$. The chance $\rv{x}$ that a student does both $A$ and $B$ correctly is $\rv{x} = \rv{a}\rv{b}$.
    \item The $\orOp$-operator: if an exercise can be solved by either doing $A$ or $B$, with each method being suitable, then we write $X = \orOp(A,B)$. In this case $\rv{x} = 1 - (1-\rv{a})(1-\rv{b}) = \rv{a} + \rv{b} - \rv{a}\rv{b}$. Note that $\rv{x} \geq \max(\rv{a},\rv{b})$ because a student can use the other method/skill to check his/her answer.
\end{itemize}
In theory an exercise can have any kind of set-up. For instance, we may have $X = \andOp(A, \orOp(A,B))$, in which case $x(\rv{a},\rv{b}) = \rv{a}^2 + \rv{a}\rv{b} - \rv{a}^2\rv{b}$. The first expression `$\andOp(A, \orOp(A,B))$' is known as the \term{exercise set-up} while the second expression `$\rv{a}^2 + \rv{a}\rv{b} - \rv{a}^2\rv{b}$' is the corresponding \term{probability polynomial}. Note that any set-up can be readily turned into a probability polynomial.

\subsection{Updating the skill success rate distributions}\label{ss:UpdatingSkillsGeneral}

Suppose that a student does exercise $X$ correctly, or alternatively he/she fails it. This then provides another data point $D_X$ that can be used to update the distributions of both $\rv{a}$ and $\rv{b}$. We consider how to update the distribution of $\rv{a}$. It works identically for $\rv{b}$, as well as any other potentially involved skill.

Our goal is to find the posterior $f_{\rv{a}}(a|D_X)$. (All previously known data $D_A$ and $D_B$ on skills $A$ and $B$ is also taken into account, but this is left out of the notation for brevity.) By marginalizing the joint distribution $f_{\rv{a},\rv{b}}(a,b|D_X)$ with respect to $\rv{b}$, followed up with Bayes' law, we find
\begin{equation}
    f_{\rv{a}}(a|D_X) = \int_0^1 f_{\rv{a},\rv{b}}(a,b|D_X) \, db = \int_0^1 \frac{p(D_X|a,b) f_{\rv{a},\rv{b}}(a,b)}{p(D_X)} \, db.
\end{equation}
Note that $p(D_X)$ is a constant, not depending on $a$, so we can leave it out if we normalize coefficients afterwards. Also note that $p(D_X|a,b)$ follows from the probability polynomial $x(a,b)$ as
\begin{equation}
    p(D_X|a,b) = \begin{cases}
    x(a,b) & {\rm on\ success,} \\
    1 - x(a,b) & {\rm on\ failing.}
    \end{cases}
\end{equation}
We generally also assume independence of $\rv{a}$ and $\rv{b}$, since they are different skills. In this case we can write $f_{\rv{a},\rv{b}}(a,b)$ as $f_{\rv{a}}(a)f_{\rv{b}}(b)$ and hence get
\begin{equation}\label{eq:PolynomialIntroduction}
    f_{\rv{a}}(a|D_X) \sim \left(\int_0^1 p(D_X|a,b) f_{\rv{b}}(b) \, db\right) f_{\rv{a}}(a) = h_{\rv{a}}(a) f_{\rv{a}}(a),
\end{equation}
where we have defined $h_{\rv{a}}(a)$ as the part between brackets in the above equation. Note that $h_{\rv{a}}(a)$ is merely, for a given $a$, the expected value of (on success) the probability polynomial $x(a,\rv{b})$ or (upon failure) the inverse $1 - x(a,\rv{b})$, with respect to all other random variables.

$h_{\rv{a}}(a)$ is a polynomial that only depends on $a$. This means we can write it as
\begin{equation}
    h_{\rv{a}}(a) = \sum_{i=0}^{n_p} k_i a^i,
\end{equation}
for some polynomial order $n_p$ and some set of constants $k_0, \ldots, k_{n_p}$. The exact value of these constants can be found from the probability polynomial with the help of~\eqref{eq:Moments}.

Before we continue, we must first put $h_{\rv{a}}(a)$ in an alternate form. We want to write $h_{\rv{a}}(a)$ as
\begin{equation}
    h_{\rv{a}}(a) = \sum_{i=0}^{n_p} k_i' \binom{n_p}{i} a^i (1-a)^{n_p - i},
\end{equation}
for another set of constants $k_0', \ldots, k_{n_p}'$. The link between these two sets of constants is
\begin{equation}
    \binom{n_p}{n_p-i} k_i' = \sum_{j=0}^{i} \binom{n_p-j}{n_p-i} k_j.
\end{equation}
Thanks to this alternate way of writing $h_{\rv{a}}(a)$, we can now update the coefficients of $f_{\rv{a}}(a)$. If this PDF used to have order $n$, with coefficients $c_i$ for $0 \leq i \leq n$, then the posterior PDF $f_{\rv{a}}(a|D_X)$ will be of order $n_* = n + n_p$. Its coefficients $c_i^*$, with $0 \leq i \leq n_*$, can be found through
\begin{equation}\label{eq:GeneralUpdateLaw}
    c_i^* = \sum_{j=\max(0,i-n)}^{\min(n_p,i)} \binom{i}{j} \binom{n_*-i}{n_p-j} c_{i-j} k_j',
\end{equation}
where afterwards coefficients must once more be normalized. Note that this method is a generalization of the update law of~\eqref{eq:BasicUpdateLaw}. It allows us to update the distributions of any skills $A$, $B$, and potentially more, based on exercises with any set-up involving these skills.

In practice this update law has some intuitive consequences. Consider the situation where a student has mostly mastered skill $A$ and is still struggling with skill $B$. If he/she then tries an exercise with set-up `$\andOp(A,B)$' and fails, then the probability theory inherent in the above method automatically searches for `blame'. In this case, the fault most likely lies in skill $B$, so that skill is more strongly penalized, while the success rate for skill $A$ is only slightly adjusted downwards. This behavior is exactly what seems sensible, showing that the system works in the desired way.

\subsection{Inferring the success rate for the exercise}\label{ss:Inference}

When considering which exercise to present to the student, it is useful to know how likely the student is to successfully complete each exercise. Given an exercise $X$ with known set-up and probability polynomial, we want to know the success rate $\rv{x}$, or more simply put the expected value $\ex[\rv{x}]$ of it.

The expected value $\ex[\rv{x}]$ is actually straightforward to find. If we once more assume that $\rv{a}$ and $\rv{b}$ are independent, then the expected value of the probability polynomial $\ex[x(\rv{a},\rv{b})]$ follows directly from the application of~\eqref{eq:Moments}.

In some cases it may also be useful to know the complete distribution of $\rv{x}$. This tells us (among others) how certain we are of the estimated success rate $\ex[\rv{x}]$. Finding the distribution of $\rv{x}$ is rather involved, mostly because it generally \textit{cannot} be described through our basis functions. However, we can define a nearly identical random variable $\rv{\hat{x}}$ whose distribution \textit{can} be described through basis functions.

We want the exercise success rate $\rv{x}$ and this new random variable $\rv{\hat{x}}$ to have as similar distributions as possible. To capture this similarity, we define the prior $f_{\rv{\hat{x}},\rv{x}}(\hat{x},x)$, identically to~\eqref{eq:JointPrior}, as
\begin{equation}
    f_{\rv{\hat{x}},\rv{x}}(\hat{x},x) = \frac{\ve{g_{n_i}}\hspace{-6pt}^T(\hat{x}) \ve{g_{n_i}}(x)}{n_i+1},
\end{equation}
for some \term{inference smoothing order} $n_i$. In practice, the order $n_i$ for this application is often chosen to be on the lower side ($n_i \approx 10$) for practical reasons we will soon see.

We want to find the posterior distribution $f_{\rv{\hat{x}}}(\hat{x}|D_{A,B})$ given all data $D_{A,B}$ on skills $A$ and $B$. Identically to how we found~\eqref{eq:SmoothingPDFs} we can find
\begin{equation}
    f_{\rv{\hat{x}}}(\hat{x}|D_{A,B}) = \int_0^1 \frac{f_{\rv{\hat{x}},\rv{x}}(\hat{x},x) f_{\rv{x}}(x|D_{A,B})}{f_{\rv{x}}(x)} \, dx.
\end{equation}
This integral can subsequently be solved and rewritten into the basis function form. This gives us a set of coefficients $c_{x,i}$ describing the posterior distribution of $\rv{\hat{x}}$. These coefficients satisfy
\begin{equation}
    c_{x,i} = \int_0^1 \int_0^1 g_{i,n_i}(x(a,b)) f_{\rv{a}}(a|D_A) f_{\rv{b}}(b|D_B) \, da \, db,
\end{equation}
where subsequent normalization must be applied, as usual. Alternatively, we can say that
\begin{equation}\label{eq:InferenceCoefficients}
    \ve{c_x} = \ex[\ve{g_{n_i}} \! (x(\rv{a},\rv{b}))].
\end{equation}
So the coefficients $c_{x,i}$ are simply the expected values of the basis functions $g_{i,n_i}$ when inserting the probability polynomial into these basis functions. This idea works identically if more than two skills $A$ and $B$ are involved, and it is known as the \term{inference step} of the algorithm.

The remaining question is how to calculate these coefficients from the coefficients of $\rv{a}$ and $\rv{b}$. Note here that each basis function $g_{i,n_i}$ is also a polynomial function. Inserting a polynomial into a polynomial function once more gives a polynomial. If $g_{i,n_i}$ does not have too large powers (that is, we keep $n_i$ reasonably small) this can be expanded through a binomial expansion. Subsequently, the expected value can be found through~\eqref{eq:Moments}.

Finally, it might be interesting to discuss the intuitive view of this new variable $\rv{\hat{x}}$: how can we interpret it? Some may argue that $\rv{x}$, which follows directly from $\rv{a}$ and $\rv{b}$ through the probability polynomial $x(\rv{a},\rv{b})$, is \textit{not} the most accurate way to describe whether the student will successfully complete exercise $X$. After all, when doing exercise $X$, the student must also recognize the steps to take, which has nothing to do with $A$ and $B$ individually, but does affect the success rate for exercise $X$. As a result, we must define a \term{true success rate} $\rv{\hat{x}}$, which takes the estimated success rate $\rv{x}$ based on the skills $A$ and $B$, and adds some uncertainty (read: smoothing of the distribution) on top of this. That's exactly what $\rv{\hat{x}}$ is.

\section{Describing relations between skills}\label{s:LinkedSkills}

The strength of the PDT algorithm lies in its ability to link related skills. A course generally does not consist of a collection of unrelated skills. The skills build up on each other in often complex ways. This can be modeled and incorporated. The complete method may seem convoluted at first, but the overview at the end (Figure~\ref{fig:InferenceOverview}) should clarify this.

\subsection{Defining the skill set-up}\label{ss:SkillSetup}

A \term{composite skill} is a skill made up of multiple smaller \term{subskills}. For instance, to calculate $2 + 3 \cdot 4$ you must first figure out the order of operations, then apply multiplication and finally apply addition. In this simple example, we could say that skill $S$ (evaluating basic expressions) has a set-up of $\andOp(A,B,C)$, with $A$, $B$ and $C$ the aforementioned subskills.

Skills have a set-up similar to exercises, but while exercises always have a \term{deterministic set-up}, only using `$\andOp$' and `$\orOp$' operations, skills may have a \term{non-deterministic set-up}. In the above example, instead of using addition, we may perhaps require subtraction half the times, resulting in a varying set-up. We therefore add the following non-deterministic operations.
\begin{itemize}
    \item The $\pickOp$-operator: from a list of subskills, select only one. For instance, if a skill first requires either $A$ or $B$ (but always only one of them) followed by $C$, then we write $S = \andOp(\pickOp(A,B),C)$. To find the probability polynomial, we may reduce `$\pickOp(A,B)$' to $\frac{1}{2}\rv{a} + \frac{1}{2}\rv{b}$. Extensions exist where multiple skills from a list are selected and/or weights are applied upon selection.
    \item The $\partOp$-operator: if a skill only requires a skill $A$ in a part $p$ of the cases, always followed by a skill $B$, then we may write $S = \andOp(\partOp(A,p),B)$. Alternatively, if sometimes we can apply either $A$ or $B$, and sometimes only $B$, we may write $S = \orOp(\partOp(A,p),B)$.
    
    For the $\partOp$-operator, the probability polynomial depends on the surrounding operator. On a surrounding $\andOp$-operator we reduce `$\partOp(A,p)$' to $1 - p(1-\rv{a})$, while on a surrounding $\orOp$-operator `$\partOp(A,p)$' becomes $p\rv{a}$.
\end{itemize}
With these extra operators, it is still always possible to turn a skill set-up into a probability polynomial. As a result, using the methods described in the inference subsection, we can always predict the distribution of the success rate of a skill $S$ based on data from its subskills, just like for an exercise $X$.

\subsection{Merging observations on skills and subskills}\label{ss:MergingDistributions}

Consider the situation where a composite skill $S$ has subskills $A$ and $B$, and where data is available on \textit{all} these skills.
\begin{itemize}
    \item First the student practices subskills $A$ and $B$, giving data $D_{A,B}$. Using inference methods -- specifically the coefficients from~\eqref{eq:InferenceCoefficients} -- we can hence infer $f_{\rv{s}}(s|D_{A,B})$.
    \item Subsequently the student practices skill $S$ directly, giving data $D_S$. Using \textit{only} this data, we can also find a distribution $f_{\rv{s}}(s|D_S)$. This is done through~\eqref{eq:BasicUpdateLaw} or more generally with~\eqref{eq:GeneralUpdateLaw}.
\end{itemize}
How can this data then be `merged' into $f_{\rv{s}}(s|D_{A,B},D_S)$? To answer this question, we must apply Bayes' law,
\begin{equation}
    f_{\rv{s}}(s|D_{A,B},D_S) = \frac{p(D_{A,B},D_S|s) f_{\rv{s}}(s)}{p(D_{A,B},D_S)}.
\end{equation}
Assuming that the observation sets $D_{A,B}$ and $D_S$ are conditionally independent given $\rv{s}$, we may reduce $p(D_{A,B},D_S|s)$ to $p(D_{A,B}|s)p(D_S|s)$. Applying Bayes' law in the reverse direction subsequently gives
\begin{equation}\label{eq:MergingPDFs}
    f_{\rv{s}}(s|D_{A,B},D_S) = \frac{p(D_{A,B})p(D_S)}{p(D_{A,B},D_S)} \frac{f_{\rv{s}}(s|D_{A,B}) f_{\rv{s}}(s|D_S)}{f_{\rv{s}}(s)}.
\end{equation}
Note that all the above factors apart from $f_{\rv{s}}(s|D_{A,B})$ and $f_{\rv{s}}(s|D_S)$ are constants. We can therefore find the posterior distribution of $\rv{s}$ by multiplying these separately inferred distributions and normalizing the result.

Let's say that $f_{\rv{s}}(s|D_{A,B})$ has order $n_{A,B}$ and coefficients $c_i^{A,B}$, and identically $f_{\rv{s}}(s|D_S)$ has order $n_S$ and coefficients $c_i^S$. The resulting posterior distribution $f_{\rv{s}}(s|D_{A,B},D_S)$ then has order $n_m = n_{A,B} + n_S$ and coefficients $c_i^m$ satisfying
\begin{equation}\label{eq:MergingCoefficients}
    c_i^m = \sum_{j=\max(0,i-n_{A,B})}^{\min(n_S,i)} \binom{i}{j} \binom{n_m - i}{n_S - j} c_{i-j}^{A,B} c_j^S,
\end{equation}
where the coefficients must be normalized afterwards. Note the similarity of the above inference law with the update law~\eqref{eq:GeneralUpdateLaw}.

The above method effectively shows how to merge two distributions of $f_{\rv{s}}(s)$, based on different, separate and conditionally independent data, together into one distribution. As a result the above method is known as \term{merging distributions}, which is also why we use a subscript/superscript $m$ here.

\subsection{Incorporating correlations between skills}\label{ss:SkillCorrelations}

The previously discussed parent/child dependencies between skills are vary common in practice. Less common, but still relevant, are correlated skills. For instance, the skill `expand brackets with numbers' with example exercise `calculate $(2+3) \cdot 4$' is generally strongly correlated with the skill `expand brackets with variables' with example exercise `expand $(a+b) \cdot c$'. Note that these skills are similar but not identical: many a mathematics teacher can confirm that students who have mastered the first skill often still struggle with the second one. Nevertheless, someone's performance on the first skill \textit{does} give a little bit of information about the expected performance on the second skill.

To model this, we consider a skill $S$ with a correlated skill $R$. We write their success rates as $\rv{s}$ and $\rv{r}$, respectively. We now want to use data $D_R$ on skill $R$ to infer the distribution of $\rv{s}$. Once more, the first step is to define a joint prior between $\rv{r}$ and $\rv{s}$. Identically to~\eqref{eq:JointPrior} we assume
\begin{equation}
    f_{\rv{r},\rv{s}}(r,s) = \frac{\ve{g_{n_c}}\hspace{-6pt}^T(r) \ve{g_{n_c}}(s)}{n_c+1},
\end{equation}
for some \term{correlation smoothing order} $n_c$. The stronger the correlation, the higher $n_c$ must be. In practice correlations are often weak, so $n_c$ does not go above a value of $10$.

Once the joint prior is defined, we notice that the problem we have here is mathematically exactly the same as inferring $\rv{a}_{k+1}$ from $\rv{a}_k$ through smoothing. The distribution $f_{\rv{s}}(s|D_R)$ is hence simply the smoothed version of $f_{\rv{r}}(r|D_R)$, using smoothing order $n_c$. The coefficients follow from~\eqref{eq:SmoothingCoefficients}.

Once $f_{\rv{s}}(s|D_R)$ has been determined, we can merge it into $f_{\rv{s}}(s|D_S)$ using the recently discussed merging methods; specifically through~\eqref{eq:MergingCoefficients}. This gives us the posterior $f_{\rv{s}}(s|D_R,D_S)$. After all, the problem is identical to when we tried to derive the posterior distribution of $\rv{s}$ based on data from multiple subskills. Additionally, in case data from subskills is \textit{also} present, we can merge that in too. The order of merging does not matter.

A more complex problem appears when there are groups of correlated skills. Imagine there are two skills $Q$ and $R$ that are both correlated with skill $S$. You could treat all these correlations individually and separately merge $f_{\rv{s}}(s|D_Q)$ and $f_{\rv{s}}(s|D_R)$ into $f_{\rv{s}}(s|D_S)$. However, this neglects correlations between skills $Q$ and $R$ which are most likely present. To take this into account, we can define a joint prior like
\begin{equation}
    f_{\rv{q},\rv{r},\rv{s}}(q,r,s) = \frac{1}{n_c+1} \sum_{i=0}^{n_c} g_{i,n_c}(q) g_{i,n_c}(r) g_{i,n_c}(s).
\end{equation}
By applying Bayes' law back and forth a few times, similar to what we did for~\eqref{eq:MergingPDFs}, we can find that the joint posterior distribution given data on $Q$ and $R$ is proportional to
\begin{equation}
    f_{\rv{q},\rv{r},\rv{s}}(q,r,s|D_Q,D_R) \sim f_{\rv{q}}(q|D_Q) f_{\rv{r}}(r|D_r) f_{\rv{q},\rv{r},\rv{s}}(q,r,s).
\end{equation}
Marginalizing over $\rv{q}$ and $\rv{r}$ then results in the posterior distribution $f_{\rv{s}}(s|D_{Q,R})$. The lengthy mathematics are omitted for reasons of brevity, but the final procedure is as follows.
\vspace{-6pt}
\begin{itemize}
    \itemsep -2pt
    \item Start with the coefficients $c_i^q$ and $c_i^r$ describing $f_{\rv{q}}(q|D_Q)$ and $f_{\rv{r}}(r|D_R)$. (Ensure that the learning effect and time decay are already taken into account.)
    \item Smooth these distributions separately with~\eqref{eq:SmoothingCoefficients} giving coefficients $c_i^{q'}$ and $c_i^{r'}$ describing $f_{\rv{s}}(s|D_Q)$ and $f_{\rv{s}}(s|D_R)$.
    \item Multiply the coefficients $c_i^{q'}$ and $c_i^{r'}$ element-wise as
    \begin{equation}\label{eq:ElementWiseCoefficients}
        c_i^{s'} = c_i^{q'} \cdot c_i^{r'}
    \end{equation}
    to find the coefficients $c_i^{s'}$ describing $f_{\rv{s}}(s|D_Q,D_R)$.
    \item Merge the distributions $f_{\rv{s}}(s|D_S)$ (described by $c_i^s$) and $f_{\rv{s}}(s|D_Q,D_R)$ (described by $c_i^{s'}$) using~\eqref{eq:MergingCoefficients}. The resulting coefficients describe $f_{\rv{s}}(s|D_Q,D_R,D_S)$ as intended.
\end{itemize}
\vspace{-6pt}
This allows us to even take into account groups of correlated skills. In practice this is hardly ever necessary: correlated duos do occur, but triplets or quadruplets are extremely rare and often indicate an inconveniently chosen course set-up.

\subsection{Overview of performance distribution tracing steps}

\begin{figure}[!ht]
\centering
\includegraphics[width=\textwidth]{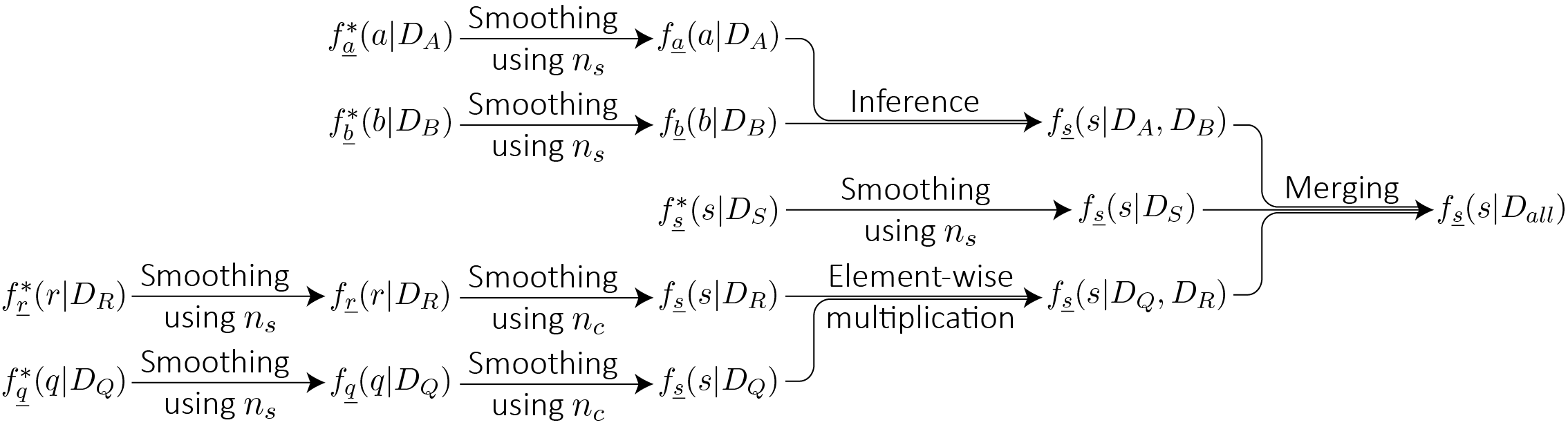}
\caption{Overview of how to incorporate data from subskills and correlated skills. The star distributions -- stored in the database -- denote the distributions \textit{after} incorporating a new observation from an exercise, but \textit{before} applying any smoothing for practice and time decay. For smoothing use~\eqref{eq:SmoothingCoefficients}, for inference use~\eqref{eq:InferenceCoefficients} and for merging use~\eqref{eq:MergingCoefficients}.}
\label{fig:InferenceOverview}
\end{figure}

We have so far seen how, given data on a skill $S$, its subskills $A, B, \ldots$ and its correlated skills $Q, R, \ldots$, the posterior distribution of $\rv{s}$ is determined. Let's create an overview. The entire procedure is summarized in Figure~\ref{fig:InferenceOverview}.

Note that the coefficients stored in the database are always the ones describing a distribution of $\rv{s}$ \textit{after} data on an exercise execution has been incorporated, but \textit{before} smoothing has been performed to take into account the learning effect (practice decay) as well as time decay. So after pulling coefficients from the database, these stored coefficients $\ve{c^*}$ are first always smoothed using~\eqref{eq:SmoothingCoefficients} with smoothing order $n_s$. Note that, as explained in Appendix~B, the value of $n_s$ depends (among others) on the time passed since the last exercise, so it is likely to have a different value for every skill.

If a skill is a composite skill, and hence has subskills, an inference step is applied using the skill set-up. This is done through~\eqref{eq:InferenceCoefficients}. Afterwards distributions are merged using~\eqref{eq:MergingCoefficients}. Similarly, if a skill is correlated with other skills, then these other skills are first smoothed with order $n_c$ using~\eqref{eq:SmoothingCoefficients}. Here $n_c$ depends on the strength of the correlation. If there is more than one correlated skill, the resulting coefficients for each of these correlated skills are multiplied element-wise using~\eqref{eq:ElementWiseCoefficients}. This distribution is then once more merged in using~\eqref{eq:MergingCoefficients}. The merging order is irrelevant.

Through this method, we can always find the most informed distribution of $\rv{s}$, incorporating as much data as is possible/sensible. And, because we always use distributions, the certainty/accuracy of said data is always correctly taken into account. In theory, if correlated skills $Q, R, \ldots$ also have subskills, or if subskills $A, B, \ldots$ also have correlated skills or further subskills, these can be taken into account as well. In practice, due to the repetitive smoothing, the tiny bit of extra information resulting from this is generally not considered worth the effort. Hence the plan of Figure~\ref{fig:InferenceOverview} suffices.

Once the PDT algorithm has determined, for every relevant skill $S$, the most well-informed performance distribution $f_{\rv{s}}(s|D_{all})$, this information can then be applied. It can be used to inform a student of his/her progress, to recommend skills to practice, or to select exercises to try out. And once a new data point becomes available -- the student tries and succeeds at or fails a certain exercise -- update law~\eqref{eq:GeneralUpdateLaw} can be used to directly update all relevant stored coefficients.

\section{Conclusions and recommendations}\label{s:Conclusions}

This paper has presented the fundamental ideas and update equations of the novel Performance Distribution Tracing algorithm, as well as shown how it can be implemented in practice. The algorithm successfully allows for live updating of skill success rate estimates, while taking into account a large variety of links between skills, and on top of that it has excellent explainability properties. This allows it to fill a gap left by other Knowledge Tracing algorithms.

For evaluation, the algorithm has been applied in practice to a group of roughly 100 students. The set-up and results of this experiment are discussed in a separate publication~\cite{StepWiseApplication}, and source code is available in Javascript through a GitHub repository~\cite{StepWiseSourceCode}. To summarize the results, students felt the skill level estimates give a realistic overview of their abilities, and having a clear overview encouraged them to practice more. Students generally followed practice recommendations based on the skill ratings, but there were exceptions. A common theme was students claiming `I only made a small calculation error, but I understand how to solve this. I should be able to move on!' But both the automatic recommendations as well as the live teachers told them, `You have to practice these basic skills more, to be able to do them well consistently.' It is fascinating to see that students disagree in exactly the same way with the algorithm recommendations as with the recommendations from live teachers. We may conclude that the goal of reaching teacher-like practice recommendations has been reached.

A useful next research step would be to statistically evaluate the algorithm, comparing its performance/prediction accuracy to similar knowledge tracing algorithms. This is very difficult, because no large-scale data sets appear with exercises that are sufficiently linked to apply PDT. On top of that, PDT uses different input data compared to other algorithm (it also incorporates student performance on exercise steps) and it gives its results in a different format (probability distributions instead of fixed numbers) which leads to the question of what performance metric would be fair. Both reasons make it very hard, if not impossible, to make a proper comparison between algorithms, leaving this as an idea for a research project on its own.

On a smaller scale, there are still many more ideas for possible extensions. Currently, we have assumed that the prior distribution for skill success rates is flat. This could be optimized, possibly skill by skill, to improve the algorithm estimates. Next to this, there is the assumption that the success rate $\rv{a}_{k+1}$ of a future skill execution has (a priori) the same mean as the success rate $\rv{a}_k$ of the previous execution. We could let go of this assumption, assuming an inherent learning effect, but this opens up the question how much this effect should be present. Using live student data this question may be answered, possibly also on a skill by skill basis or even on a per-student basis, detecting quick learners from those who pick things up more slowly.

A more interesting extension for practical applications takes into account collaborations between students. Often students practice together. If two students, each with their own individual data $\rv{s}_1$ and $\rv{s}_2$, do an exercise together, how can we predict the success rate $\rv{s}_{1,2}$ of them working together? And how should each student's coefficients subsequently be updated? This is a highly relevant extension, since collaborative practice is very strongly encouraged at many universities, and it may help digital practice support tools to bring students together.

\vspace{\fill}\pagebreak

\appendix
\renewcommand{\theequation}{A\arabic{equation}}
\setcounter{equation}{0}
\renewcommand{\thesection}{\Alph{subsection}}
\setcounter{section}{0}


\section{Appendix A. Mathematical derivation of coefficient equations}\label{s:CoefficientDerivation}

This appendix has as goal to show how an expression with PDFs like~\eqref{eq:SmoothingPDFs} can be turned into an expression for coefficients like~\eqref{eq:SmoothingCoefficients}. We will go through the full derivation, first isolating coefficients and then solving the integral inside their expression.

\subsection{Isolating the coefficients}

Our first task is isolating the coefficients. We start at~\eqref{eq:SmoothingPDFs}, for easy reading repeated as
\begin{equation}\label{eq:SmoothingPDFsStartingPoint}
    f_{\rv{a}_{k+1}}(a_{k+1}|D_k) = \int_0^1 \frac{f_{\rv{a}_k,\rv{a}_{k+1}}(a_k,a_{k+1}) f_{\rv{a}_k}(a_k|D_k)}{f_{\rv{a}_k}(a_k)} \, da_k.
\end{equation}
In this expression we have the joint prior $f_{\rv{a}_k,\rv{a}_{k+1}}(a_k,a_{k+1})$ defined by~\eqref{eq:JointPrior}, or written in its sum notation as
\begin{equation}
    f_{\rv{a}_k,\rv{a}_{k+1}}(a_k,a_{k+1}) = \frac{1}{n_s + 1} \sum_{i=0}^{n_s} g_{i,n_s}(a_k) g_{i,n_s}(a_{k+1}).
\end{equation}
Similarly, the posterior distribution of $\rv{a}_k$, described by $f_{\rv{a}_k}(a_k|D_k)$, follows from~\eqref{eq:PDFWithCoefficientsAndBasisFunctions} as
\begin{equation}
    f_{\rv{a}_k}(a_k|D_k) = \sum_{i=0}^{n_*} c_{k,i}^* g_{i,n_*}(a_k),
\end{equation}
where $c_{k,i}^*$ are the coefficients describing $\rv{a}_k$, after incorporating recent observations. These updated coefficients have been calculated through~\eqref{eq:BasicUpdateLaw} or alternatively~\eqref{eq:GeneralUpdateLaw}. Finally, we have the prior $f_{\rv{a}_k}(a_k)$ which equals $1$ and can hence be ignored.

Inserting all the above expressions into~\eqref{eq:SmoothingPDFsStartingPoint} turns it into
\begin{equation}
    \int_0^1 \frac{1}{n_s+1} \left(\sum_{i=0}^{n_s} g_{i,n_s}(a_k) g_{i,n_s}(a_{k+1})\right) \left(\sum_{i=0}^{n_*} c_{k,i}^* g_{i,n_*}(a_k)\right) da_k.
\end{equation}
We can rearrange the above to write it as
\begin{equation}
    \sum_{i=0}^{n_s} \left(\frac{1}{n_s+1} \int_0^1 \left(\sum_{j=0}^{n_*} c_{k,j}^* g_{j,n_*}(a_k)\right) g_{i,n_s}(a_k) \, da_k \right) g_{i,n_s}(a_{k+1}).
\end{equation}
Comparing this with the standard basis function notation~\eqref{eq:PDFWithCoefficientsAndBasisFunctions}, we can directly see that the coefficients $c_{k+1,i}$ describing $f_{\rv{a}_{k+1}}(a_{k+1}|D_k)$ equal
\begin{equation}\label{eq:IsolatedCoefficient}
    c_{k+1,i} = \frac{1}{n_s+1} \int_0^1 \sum_{j=0}^{n_*} c_{k,j}^* g_{j,n_*}(a_k) g_{i,n_s}(a_k) \, da_k.
\end{equation}
The coefficients have hence been isolated. What remains is solving the above integral.

\subsection{Solving the integral}

Consider~\eqref{eq:IsolatedCoefficient}. Note that, if we normalize the coefficients $c_{k+1,i}$ afterwards, we can safely ignore constant multiplications: factors not depending on $i$. Hence $\frac{1}{n_s+1}$ drops out. If we expand the basis functions using their definition~\eqref{eq:BasisFunctionDefinition}, and ignore the constant factors $(n+1)$ within, we can write the above expression for $c_{k+1,i}$ as
\begin{equation}
    c_{k+1,i} = \int_0^1 \sum_{j=0}^{n_*} c_{k,j}^* \binom{n_*}{j} a_k^j (1-a_k)^{n_*-j} \binom{n_s}{i} a_k^i (1-a_k)^{n_s-i} \, da_k.
\end{equation}
Rearranging the above turns it into
\begin{equation}\label{eq:SolvingCoefficientsIntermediateStep}
    c_{k+1,i} = \sum_{j=0}^{n_*} c_{k,j}^* \binom{n_s}{i} \binom{n_*}{j} \int_0^1 a_k^{i+j} (1-a_k)^{n_*+n_s-i-j} \, da_k.
\end{equation}
Note that in general, for any integer values of $i$ and $n$ with $0 \leq i \leq n$, it holds that
\begin{equation}\label{eq:BasisFunctionIntegralIsOne}
    \int_0^1 g_{i,n}(x) \, dx = \int_0^1 (n+1) \binom{n}{i} x^i (1-x)^{n-i} \, dx = 1.
\end{equation}
After all, integrating over the PDF of a beta distribution results in one. Using the above relation, we can solve the integral from~\eqref{eq:SolvingCoefficientsIntermediateStep}. Ignoring constant multiplications as well, we find
\begin{equation}
    c_{k+1,i} = \sum_{j=0}^{n_*} \frac{\binom{n_s}{i} \binom{n_*}{j}}{\binom{n_*+n_s}{i+j}} c_{k,j}^*.
\end{equation}
By applying the definition of the binomial, the above can be expanded into
\begin{equation}
    c_{k+1,i} = \sum_{j=0}^{n_*} \frac{\frac{n_s!}{i!(n_s-i)!} \frac{n_*!}{j!(n_*-j)!}}{ \frac{(n_*+n_s)!}{(i+j)!(n_*+n_s-i-j)!}} c_{k,j}^* = \sum_{j=0}^{n_*} \frac{\frac{(i+j)!}{i!j!} \frac{(n_*+n_s-i-j)!}{(n_*+n_s)!(i+j)!}}{ \frac{(n_*+n_s)!}{n_*!n_s!}} c_{k,j}^*.
\end{equation}
The denominator here equals $\binom{n_*+n_s}{n_*}$, which is a constant, so it can be ignored. We remain with
\begin{equation}\label{eq:SmoothingCoefficientsAsDerived}
    c_{k+1,i} = \sum_{j=0}^{n_*} \binom{i+j}{i} \binom{n_* + n_s - i - j}{n_* - j} c_{k,j}^*,
\end{equation}
which is the final result~\eqref{eq:SmoothingCoefficients} that we wanted to get. Note that in theory it is possible to take into account all constants in the derivation of $c_{k+1,i}$ too, but in practice it is much easier to ignore constants and simply normalize coefficients afterwards. In practice, this is already done for numerical reasons anyway.

\section{Appendix B. Choosing and applying the smoothing order $n_s$}\label{s:SmoothingOrderChoice}

In the main body of this paper it has been discussed that, to incorporate the learning effect and practice decay, the distribution of the success rate $\rv{a}$ should be smoothed after every exercise. This is done with some smoothing order $n_s$. The question remains: how should $n_s$ be chosen? This is not a mathematical matter, but one of settings and preferences.

\subsection{Links between original and smoothed distributions}

Consider a skill $A$. Let's write the success rate \textit{before} smoothing as $\rv{a}_k$ and the success rate \textit{after} smoothing as $\rv{a}_{k+1}$. In this case, based on~\eqref{eq:SmoothingCoefficients}, a link can be determined between their expected values $\ex[\rv{a}_k]$ and $\ex[\rv{a}_{k+1}]$. To be precise, this link satisfies
\begin{equation}
    \left(\ex[\rv{a}_{k+1}] - \frac{1}{2}\right) = \frac{n_s}{n_s + 2} \left(\ex[\rv{a}_k] - \frac{1}{2}\right).
\end{equation}
In words, the smoothed success rate $\rv{a}_{k+1}$ always has a mean closer to $1/2$ than the original success rate $\rv{a}_k$. To be precise, the reduction factor that is applied is
\begin{equation}\label{eq:DecayRatio}
r = \frac{n_s}{n_s + 2}.
\end{equation}
We call $r$ the \term{decay ratio}. If it equals $1$ there is no smoothing, while a decay ratio of $0$ means everything is forgotten. There are now two questions: which decay ratio $r$ is appropriate? And how do we turn it into a smoothing order $n_s$?

\subsection{Choosing a decay ratio}

Smoothing should take into account both the learning effect and time decay. For time decay an exponential decay seems appropriate. Something like
\begin{equation}
    r = \left(\frac{1}{2}\right)^{t/t_{1/2}},
\end{equation}
where $t$ is the time since the last exercise and $t_{1/2}$ is the time after which the student has `lost' half of its skills. The latter is generally set to a year.

For practice decay, we could introduce another decay factor, but we do not do so. Instead, slightly more intuitively, we use the concept of \term{equivalent time}. We say that practicing one more exercise is equivalent to a time $t_e$ (for instance two months) of not practicing. We call $t_e$ the \term{equivalent inactive time}. This gives a decay ratio of
\begin{equation}
    r = \left(\frac{1}{2}\right)^{(t + t_e)/t_{1/2}}.
\end{equation}
However, we still go one step further. The first time a student does an exercise, the learning effect is still strong. It's the student's first time, so he/she is expected to learn a lot from it. However, after ten exercises or so, this effect is a lot smaller, and exercise fifty is unlikely to still have much of a learning effect at all. To incorporate that, we reduce the equivalent inactive time $t_e$ based on the amount of practice a student has had. If the student has already practiced the skill $n$ times, then we define
\begin{equation}
    t_e = t_{e,0} \left(\frac{1}{2}\right)^{n/n_{1/2}}.
\end{equation}
Here $t_{e,0}$ is the \term{initial equivalent inactive time}, set to two months, and $n_{1/2}$ is the number of times a student must practice a skill to get half the learning effect. Generally we set $n_{1/2} = 8$, although ideally this is larger for smaller (more simple) skills and smaller for larger (more complex) skills.

Using the above equations, we can always calculate an appropriate decay ratio $r$. This then needs to be applied.

\subsection{Applying a decay ratio}

To apply a given decay ratio $r$, we could simply choose a smoothing order. From~\eqref{eq:DecayRatio} you would expect it to equal
\begin{equation}
    n_s = \frac{2r}{1-r}.
\end{equation}
However, the smoothing order $n_s$ must be an integer, which complicates matters. A rough solution would be to round $n_s$ to the nearest integer value and apply that, but this results in discontinuities. Perhaps, from one day to the next, a student's score suddenly jumps. That is not desirable and may result in confusion/frustration for the end-user.

A cleaner and more continuous solution would be to write $r = r_1 \cdot r_2 \cdot r_3 \cdot \ldots$, for which each individual decay subratio $r_i$ does result in an integer smoothing order $n_{s,i} = \frac{2r_i}{1-r_i}$. We can then apply smoothing multiple times. Choosing the subratios $r_1, r_2, \ldots$ is done by applying the following steps for $i = 1, 2, \ldots$.
\vspace{-6pt}
\begin{itemize}
    \itemsep -2pt
    \item Choose the smoothing order $n_{s,i} = \lceil \frac{2r}{1-r} \rceil$.
    \item Calculate $r_i = n_{s,i}/(n_{s,i}+2)$ accordingly.
    \item Update the remaining decay ratio $r$ using $r \leftarrow r/r_i$.
    \item If $n_{s,i} > n_{s,max}$ then stop: ignore $r_i$ and further.
\end{itemize}
\vspace{-6pt}
In practice, for numerical/practical reasons, usually $n_{s,max}$ is set around $100$ or $120$ or so. This also ensures that we often get at most three subratios $r_1, r_2, r_3$, or in very rare cases four.

It must be noted here that the subratios $r_1, r_2, \ldots$ that follow from the above steps are in decreasing size -- largest to smallest -- meaning their corresponding smoothing orders $n_{s,1}, n_{s,2}, \ldots$ are in increasing size: smallest to largest. Mathematically the order in which these smoothing orders are applied does not matter, but in practice it is wisest to apply them largest to smallest, ending with $n_{s,1}$. This ensures that the final coefficient vector $\ve{c}$ we wind up with has a relatively small order $n_{s,1}$. This reduces the number of coefficients we need to work with later on, slightly speeding up further computations.

\newpage
\bibliographystyle{apacite} 
\bibliography{References} 


%




\end{document}